# Time-Critical Action: Representations and Application

**Eric Horvitz**
Microsoft Research
One Microsoft Way
Redmond, WA 98052
**horvitz@microsoft.com**

**Adam Seiver**
Critical Care Service
Department of Surgery
Stanford Medical School
Stanford, CA 94305
**seiver@smi.stanford.edu**

## Abstract

We review the problem of time-critical action and discuss a reformulation that shifts knowledge acquisition from the assessment of complex temporal probabilistic dependencies to the direct assessment of time-dependent utilities over key outcomes of interest. We dwell on a class of decision problems characterized by the centrality of diagnosing and reacting in a timely manner to pathological processes. We motivate key ideas in the context of trauma-care triage and transportation decisions.

## 1 Introduction

The nature and timing of actions by decision makers is often influenced by the pressures generated by time-dependent processes. We will examine the opportunity to minimize effort expended on modeling complex probabilistic dependencies over time by shifting attention to the definition and direct assessment of states that summarize important outcomes at future times. After touching on the general problem of planning and action in time-critical contexts, we introduce a set of time-critical decision problems and representational simplifications with the goal of bypassing the handling of complex probabilistic dependencies over time. Finally, we will highlight the ideas in the context of the real-world domain of time-critical medicine.

## 2 Representations of Time-Dependent Processes

Let us first start with a simple atemporal decision problem. Figure 1 displays an influence diagram for a simple, one-shot decision problem. Such formulations typically avoid making temporal relationships explicit. As indicated by the dependencies, the utility of the outcome is a function of one or more hypothesis variables $H$ that represent states of a system of interest. Such states are typically not directly observable. However, we can make inferences about the probability distribution over states of the hypothesis nodes by considering patterns of evidence that are influenced by the system. As indicated by the influence diagram, the hypothesis variable influences the values of a set of observables, labeled as variables $E$. Some subset of these evidential distinctions may be observed before a decision is made, as indicated by the information arcs extending from several of the observables to the decision node.

As indicated in the influence diagram, the utility of action is influenced by the action taken as well as by the state of the system or world. Although we employ a single system variable in the influence diagram to represent critical aspects about the state of the world, the more general case of multiple hypotheses can be represented by adding additional hypothesis nodes.

### 2.1 Temporal Processes and Sequences of Actions

Let us now consider the problem of multiple actions and outcomes taken over time. In the general case, we must consider the expected utility of outcomes influenced by different sequences of actions. Figure 2 displays an influence diagram for a set of actions over time. Rather than considering actions and value in an atemporal or only implicitly temporal context, we now index actions, observations, and states of the world by time. As indicated in Figure 2, we seek to generate an ideal sequence of actions or plan over time that maximizes expected utility, given background knowledge and a set of observations seen over time. In the general case, identifying such an ideal policy, where actions include decisions to gather information, is the challenging *planning under uncertainty* problem that has been receiving increasing attention by the UAI community [3, 4, 5, 16].

Let us explore the model in Figure 2 more closely. Only some of the arcs, showing a subset of potential dependencies among variables over time, are high-lighted in this figure. We represent in the model the notion that the utility of outcomes is influenced by the state of a system or the world at some time in the future. Also, this prototypical model represents the common situation that the states of the world at time $t$ often influence states of the world at time $t+1$, capturing the notion that the current state of a system has influence on future states. Many processes that show fluidity of change in variables of a system can be represented by this type of Markov dependency. Laying at the foundation of such evolving dependent physical processes is a combination of persistence in the overall structural fabric of objects and of ongoing temporal influences; we typically take for granted such stability and temporal influence in a large battery of physical processes that we encounter daily.

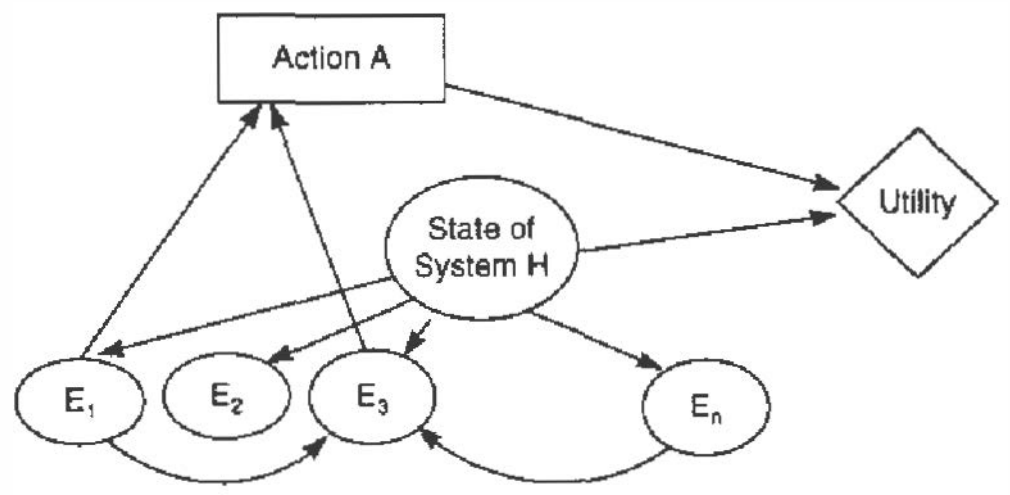

Figure 1: A simple decision model. Information arcs pointing into the decision node represent information known before action is taken. We use a single hypothesis variable here to represent one or more significant variables about the state of a system or world.

As indicated by the dependencies between actions and states, actions can directly influence states at increasingly later periods of time depending on the temporal properties of the action and states of the world. Also, observations that are available in the present moment, $t_o$, may have been caused or influenced directly by states existing at previous times. Such temporal delays in the response to action and in the emitting of evidence can lead to a variety of challenging modeling problems. For example, in many cases, we know that the evidence we are seeing in the present are caused by states of the system in earlier periods. As an example, the modeling and assessment for the Vista decision-theoretic monitoring application that has been used at NASA Mission Control Center [7], took into consideration the expected delay induced by the process of processing and beaming to earth telemetry from the Space Shuttle.

In work on Markov decision processes, utility is often computed for sequences of states as a function of rewards assigned to individual states reached by a decision making agent [12, 14, 17]. In a variety of decision contexts it can be difficult to structure the problem to represent value as rewards on intermediate states. Rather, it can be most appropriate and efficient to assess the utility over specific critical states as the value of the future life lottery associated with these outcomes. Such outcomes are often defined in terms of important results of actions taken in response to a set of critical challenges.

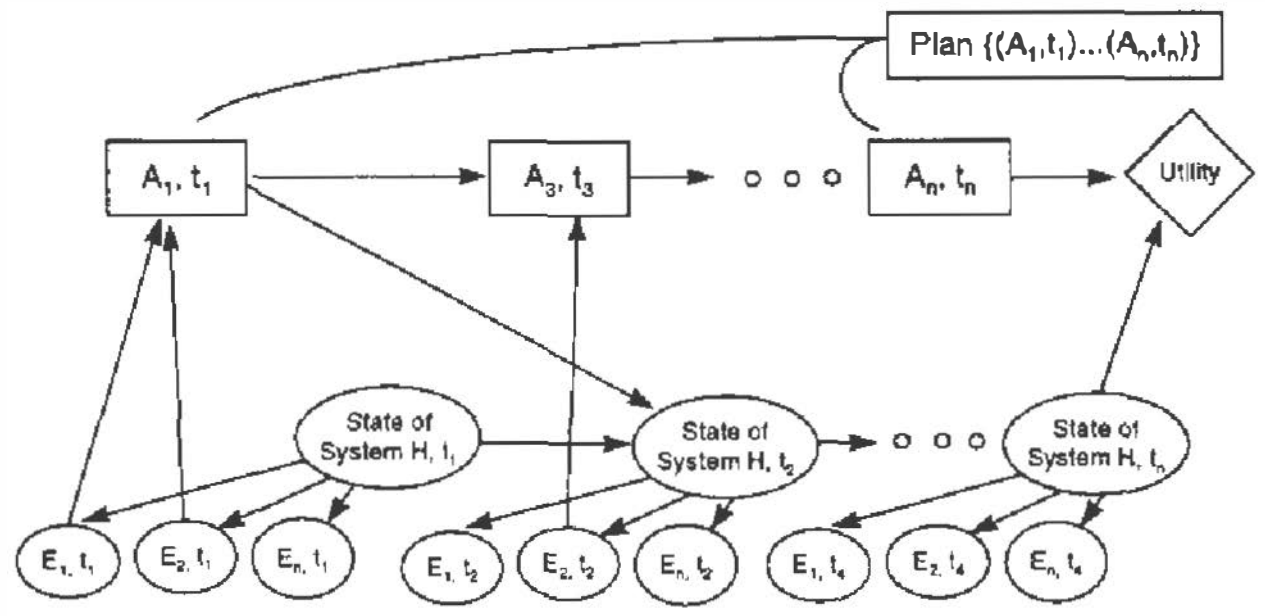

Figure 2: Representing time and sequences of actions. In this richer model, knowledge about time-dependent processes is encoded in the form of variables representing the evolution of the system over time. A subset of many of the typical dependencies are displayed.

## 3 Action and Cost in Time-Critical Contexts

Timely action is often critical in facing real world challenges. Time-critical contexts are situations where the expected value of an outcome is diminished as some function of the delay of taking one or more actions available to a decision maker or decision making system. Real-world, time-critical decision problems are typically cast in the context of an acute challenge or opportunity which heralds the initiation of a decision context. Challenges include a variety of processes and influences that can threaten valued stability or equilibria in a system, or that reduce quantities of a valued commodity. Opportunities include the development of situations that are associated with new possibilities or means to achieve a desired state of affairs.

Representations of time-dependent probability and utility have been discussed in several communities. There has been growing interest in time-dependent utility in planning [1, 5] and automated decision making [9]. Decision analysts working in such areas as medical decision making have considered time-dependent probability and utility in building models for consultation on time-pressured decisions (e.g., see [15]). A variety of formal models of urgency and their relationships to the value of computation is explored in [11]. The consideration of Bayesian models containing explicit temporal dependencies has been explored in detail by Dagum, et al. [2]. More recently, researchers have been exploring a variety of methods for efficiently solving such temporal Bayesian models [13]. Finally, there is a rich literature exploring the consideration of time-dependent processes in partially observable Markov decision processes.

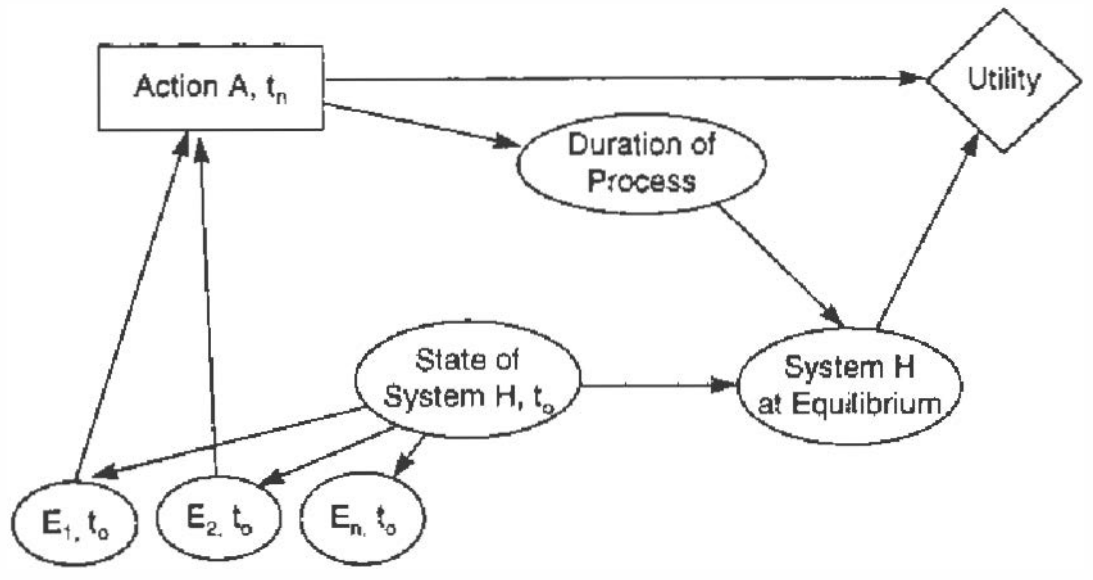

Figure 3: A formulation of a pathological-process problem. For this indicator model, the utilities of states reached at system equilibrium are assessed. The action taken influences the duration of a noxious process on the system equilibrium.

Our investigation of simplifications for time-critical reasoning problems has been influenced by our experience with modeling and inference in the area of high-stakes, time-critical arenas including trauma care and aerospace [8, 6, 7]. In particular, we have explored simplifications for specific classes of problems, centering on the definition of outcomes and on the development of techniques for summarizing outcomes in terms of time-dependent probability and utility.

## 3.1 Characterizing Time-Critical Processes

Effective automated inference about ideal action may require solving planning problems that explicitly manipulate complex temporal dependencies and that consider large sequences of action, observations, and states of the world. However, in may often be appropriate to assume simpler models and to expend effort on the direct assessment of critical, summarizing outcomes.

Figure 3 shows an influence diagram for a class of time-critical decision problem we refer to as *pathological-process problems*. The version of the problem represented by the influence diagram in the figure captures extremely time-critical situations such as trauma-care transportation decision support and the propulsion-system problem at the NASA Mission Control Center.

Pathological-process problems are typically associated with well-characterized sequences of states or processes that occur over time. Also, the expected utility of outcomes of interest in these settings is often highly dependent on an initial burst of potentially corrective actions; significant losses are associated with delay in taking appropriate action. In many time-critical pathological-process contexts, an initial pattern of evidence is observed and a decision must be made about an action. The outcome is relatively insensitive to resources aimed at interleaving information collection and action. An example of a pathological process represented in the NASA Vista system is the commencement of a *propellant failure* which changes the mixture of fuel and oxidizer, raising the temperature of the exhaust plume to a level that can destroy key engine components.

We can gracefully introduce additional complexity into simple pathological process models to represent critical context, as well as intermediate actions, observations, and outcomes–especially for less time-critical situation. Pathological processes can often be modeled as broad classes of sequences of state transitions that are parameterized by specific contexts representing specializations in the system and process. Furthermore, a variety of control actions can modulate pathological processes. Richer decision models can represent how actions can diminish the effect of a pathological process; we may wish to represent with dependencies the influence of a variety of actions on the state of the evolving system facing a dangerous erroneous process. As an example, consider the case of a patient that begins to hemorrhage a large quantity of blood into the abdomen following a blunt a injury. The utility of the outcome at equilibrium depends on the duration of the process as well as on the interventional actions, such as administering some quantity of fluids to the patient. We will return to the case of introducing intermediate actions that modulate a pathological process for the trauma transportation decisions.

## 3.2 From Intermediate States to Key Outcomes

In the context of time-critical pathological processes, we can often reformulate problems of the form represented in Figure 2 into the simpler problem displayed in Figure 3 by abstracting away details of the multi-stage probabilistic dependency model and reformulating the problem into states or outcomes that summarize complex transitions over time.

Identifying and assessing the outcomes of interest are important in formulating pathological process problems. In one approach, we assess the probability of key states or the utility over critical outcomes at some *specific* future period. In such a *fixed horizon* model, we define a process and directly assess the probability of states or the utility of outcomes $n$ periods from the current moment or from the time a challenge is noted.

Another approach is to model and assess the probability or utility of states in terms of some indicator or sentry event. When modeling outcomes with such *indicator models*, we assess the utility on the outcome representing the future life history at the time the sentry event occurs conditioned on the state of the system at that time. An example of this type of assessment is the utility over outcomes defined as states of a system achieved when an some notion of *equilibrium* is reached, following a destabilizing challenge. Such an assessment is valuable for such realms as controlling processes that may go awry but that will eventually stabilize in some stable configuration or sum-mary outcome. For example, the utility associated with outcomes of pathological processes in propulsion systems that are used to guide the Space Shuttle can be assessed as the value of the ultimate short-term outcome, following a set of firing and system control actions, which lead to a new trajectory of the Shuttle and to potential damage to systems that can provide future propulsion actions [7].

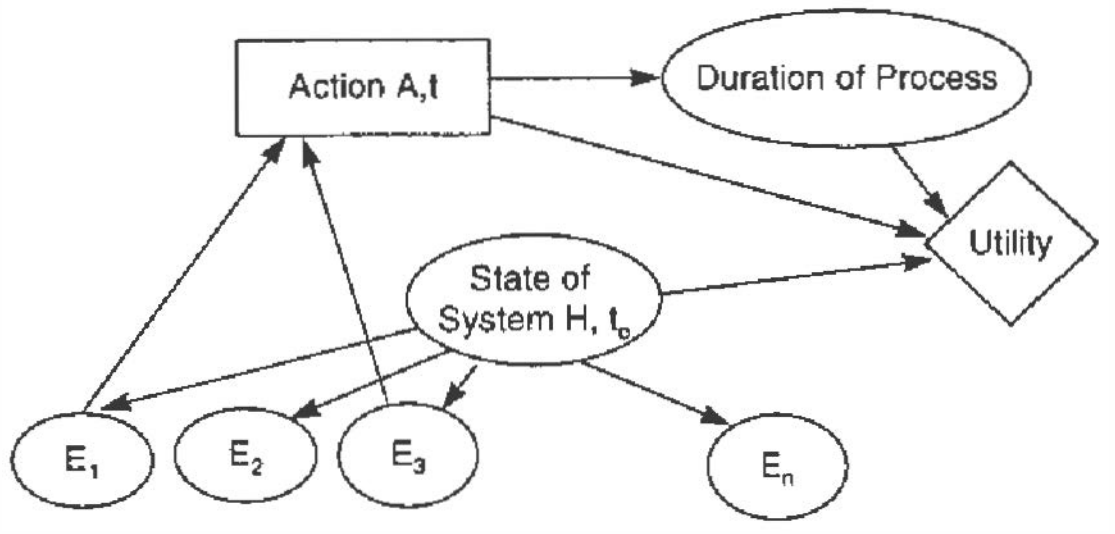

Figure 4: Direct assessment of time-dependent utility. In some cases, it can be useful to formulate the task of characterizing time-dependence as the direct assessment of time-dependent utility on outcomes defined by the intial state, corrective action, and duration of a pathological process.

For the fixed horizon or the indicator models, it can be useful to assess outcomes directly in terms of the utilities of the states of interest. As highlighted by the model in Figure 4, rather than assessing and encoding potentially complex intermediate time-dependent changes in the probability of a system given different actions and duration of the pathological process, we summarize outcomes by assessing a time-dependent utility, $u(A, H, t)$ as a function of the action taken, the state of the system, and the duration $t$ of the pathological process. Such direct assessment was explored in building the NASA Vista system for providing decision support for ground controllers about the best way to react to complex telemetry. Time-dependent utility was assessed as parameterized functions of the state, action, and duration of the process. We found that performing such a reformulation of the assessment task can ease the burden of modeling and assessing details of time-dependent probabilities.

# 4 Expected Cost of Delayed Action

A useful measure for modeling and assessing outcomes is the expected cost of delayed action. The principle of maximum expected utility, dictated by the axioms of utility, specifies that a decision maker should take actions with the greatest expected utility. In time-critical contexts, the utility of the outcomes of action are a function of the actions and the time at which they are taken. We shall use $u(A_i, H_j, t)$ to represent the time-dependent utility of outcomes of actions $A$ in the context of time-dependent processes $H$. We can view $A$ as a single action or some sequence of actions initiated at time $t$. The best action $A^*(t)$ at time $t$ is,

$$A^*(t) = \arg\max_A \sum_j u(A_i, H_j, t) p(H_j|E, \xi) \quad (1)$$

where $p(H_j|E, \xi)$ is the probability of processes $H$, given evidence $E$ and background state of knowledge $\xi$. Note that, as the system we are monitoring and attempting to control is evolving in time, the best action at time $t$, $A^*(t)$, is not necessarily equal to the best action $A^*(t')$ at a later time $t'$.

In assessing time-dependent utility, we can often summarize the variation in the utility over a large set of outcomes as a function of the duration of a process with parameterized functions that capture the basic structure of the cost of time. Prototypical cost functions include urgency, deadline, uncertain deadline, and several variants of cost built as combinations of these cost functions [10]. Such prototypical cost functions are recurrent in many real-world applications and arise in common interactions such as lost opportunity and competition for limited resources. The urgency context refers to the class of utility functions that assign cost or diminishment of value as some monotonically increasing function of delay. The deadline context refers to cases where cost is zero or negligible until some value of delay is reached, at which time a significant cost is incurred. The uncertain deadline is the common situation of facing a uncertain deadline, represented as a probability distribution over a deadline.

## 4.1 Characterizing Losses with Delay

We can characterize the cost of delay in a decision context with the expected cost of delayed action (ECDA). The ECDA is the difference in expected value of taking immediate ideal action, at time $t_o$, and delaying ideal corrective action until time $t$,

$$\text{ECDA} = \max_A \sum_j u(A_i, H_j, t_o) p(H_j|E, \xi) - \max_A \sum_j u(A_i, H_j, t) p(H_j|E, \xi) \quad (2)$$

We can compute the ECDA for any two moments in time, $t_o$ and $t$. We typically take $t_o$ to be the present moment and compute the ECDA for delays in action until time $t$.

Variants of the ECDA measure have been used in earlier decision support systems. The measure was first developed for prioritizing the display of information in the Vista-I system [8]. Beyond using the cost of delayed action for prioritizing the display of possible faults, Vista research also experimented with the display of an overall measure of criticality of a decision context, based on the rate at which the expected value of the best action is diminishing with time.

### 4.2 Modeling Uncertainty about the Initiation of the Process

Several specifications on ECDA can be useful. We use *comprehensive ECDA* to refer to the ECDA measure of loss between the time a faulty process begins and the time that corrective action is taken. However, we may not know how long a faulty process has been underway when it is diagnosed, or may be uncertain about the time a faulty process began. In such cases, even an immediate response is associated with some comprehensive ECDA.

Handling the uncertainty in the initiation of a pathological process is especially important in cases where the losses associated with delayed action are nonlinear in the duration of a process. In such cases, computing the ECDA for additional delay depends on the amount of time that has transpired since a process began. Thus, we must consider a probability distribution over the duration of the process, $d$,

$$\text{ECDA} = \sum_d p(t_o = d|E,\xi)[\max_A \sum_j p(H_j|E,\xi)u(A_i,H_j,d) - \max_A \sum_j p(H_j|E,\xi)u(A_i,H_j,t)] \quad (3)$$

### 4.3 Default State of Control

The notion of delayed action in the context of a pathological process $H$, as captured in Equation 2, leaves implicit fundamental intuitions about default states of control. Delayed action is more explicitly characterized in terms of a sequence of states of control. For many complex systems, failures or pathologies requiring time-critical corrective action lead to a prototypical sequence of states over time. Thus, to be more precise, we must define the cost of delay in terms of a corrective action or set of actions coming after some period of a *default state of control*.

We can make a default state of control explicit in ECDA computations by including a default *control context*, $C$ in the utility of action $u(A_i|C[t_o,t],H_j,t)$, where $A|C[t_o,t]$ refers to corrective action $A$ being taken at time $t$ in the context of the persistence of a default state of control active between time $t_o$ and time $t$. The control context can can represent simple states of control or more complex, evolving control dynamics. Control dynamics are defined by the default manner a system will react in the context of pathological process. For simplicity, we shall typically drop the explicit specification of the default control context, keeping in mind that it can be important to re-introduce the specification on the default state of control.

### 4.4 Considering Suboptimal Future Actions

In the real world we cannot always assume that the best action will be taken. The ECDA is an upper bound on the cost of delay. Unfortunately, suboptimal decisions may ultimately be taken in an attempt to control a faulty process. We can integrate into the measure of the cost of delay consideration that such suboptimal decisions will take place. We term such a revised measure, the *expected cost of delay and misdiagnosis* (ECDM). To compute the ECDM, we must consider the probability distribution over the actions that will be undertaken after delay,

$$\text{ECDM} = \max_A \sum_j u(A_i,H_j,t_o)p(H_j|E,\xi) - \sum_i p(A_i|E,\xi)\sum_j u(A_i,H_j,t)p(H_j|E,\xi) \quad (4)$$

## 5 Example: Triage and Transportation in Trauma Care

We will now highlight several key ideas in the context of time-critical decisions about the prioritization of transportation and treatment for victims of trauma. Trauma of various types and severities leads to time-critical medical emergencies via the destabilization of the victim's physiological machinery. Beyond the case of trauma in a single patient, natural and man-made disasters can lead to situations where multiple casualties may need medical attention at a trauma facility. We have explored the use of time-critical pathological process models for supporting time-critical decisions about trauma patient triage and transportation.

### 5.1 Modeling Urgency at a Trauma Scene

One of the problems with effective triage of victims at a trauma scene is that skilled trauma experts are typically unavailable to provide assistance with diagnosis and triage. As part of an effort to build an experimental system for assisting nonexperts with trauma-care triage, we worked with expert trauma-care surgeons on the construction of Bayesian models for diagnosing pathological processes from context, signs, and symptoms that could be easily interpreted by paramedics at a trauma site. As part of this work, we defined key classes of physiological syndromes, representing pathological processes arising from time-critical pathophysiologies. We assessed time-dependent utilities for the classes as a function of delays in transporting patients to a trauma facility. The diagnostic model and time-dependent utilities enables us to compute measures of ECDA that can be used to support patient transportation decisions.

A Bayesian network for generating a probability distribution over pathological processes is displayed in Fig-ure 5. This model was simplified by defining hypotheses of interest as a set of mutually exclusive and exhaustive primary physiological problems with trauma patients. The *primary* physiological problem is defined as the most critical pathological process facing the patient. Although the notion of primary problem was useful in simplifying the model, this formulation does not represent the potentially greater urgencies associated with the coexistence of multiple physiological problems. Findings in the model include variables that capture a variety of temporal notions that are useful for discriminating pathological processes such as observation of *increasing chest retractions* or *decreasing perfusion* over specific quantities of time.

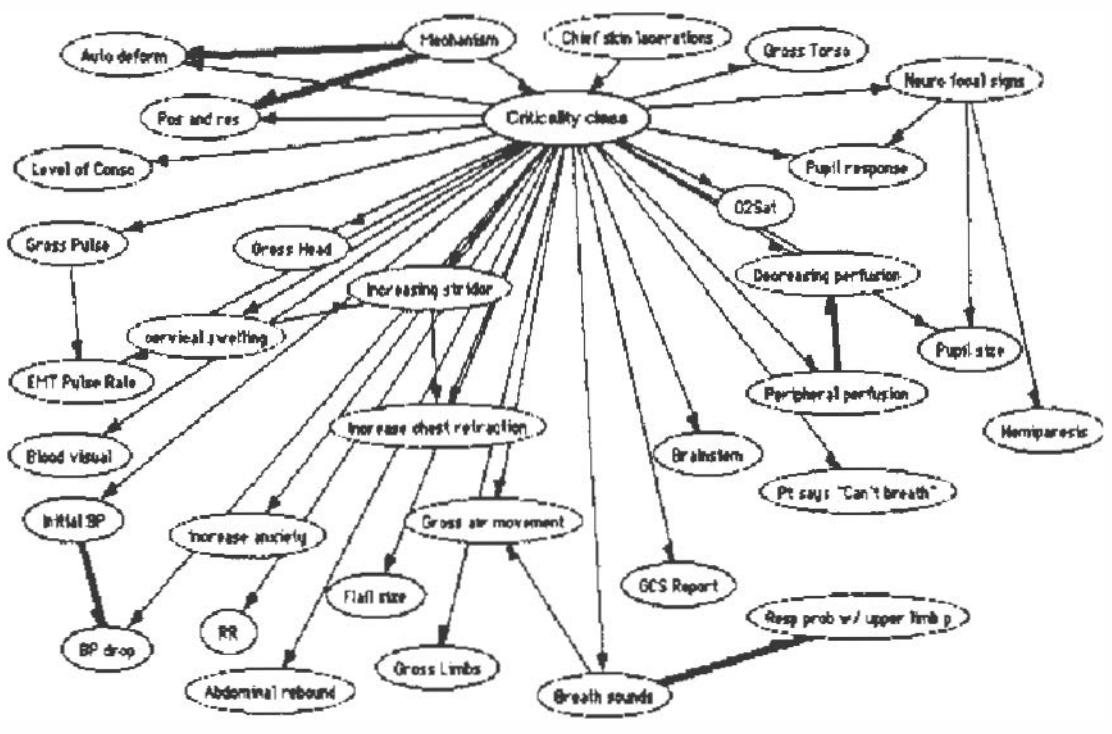

Figure 5: A Bayesian network constructed to infer the probability of key physiologic processes associated with different urgencies from signs and symptoms observed at a trauma scene.

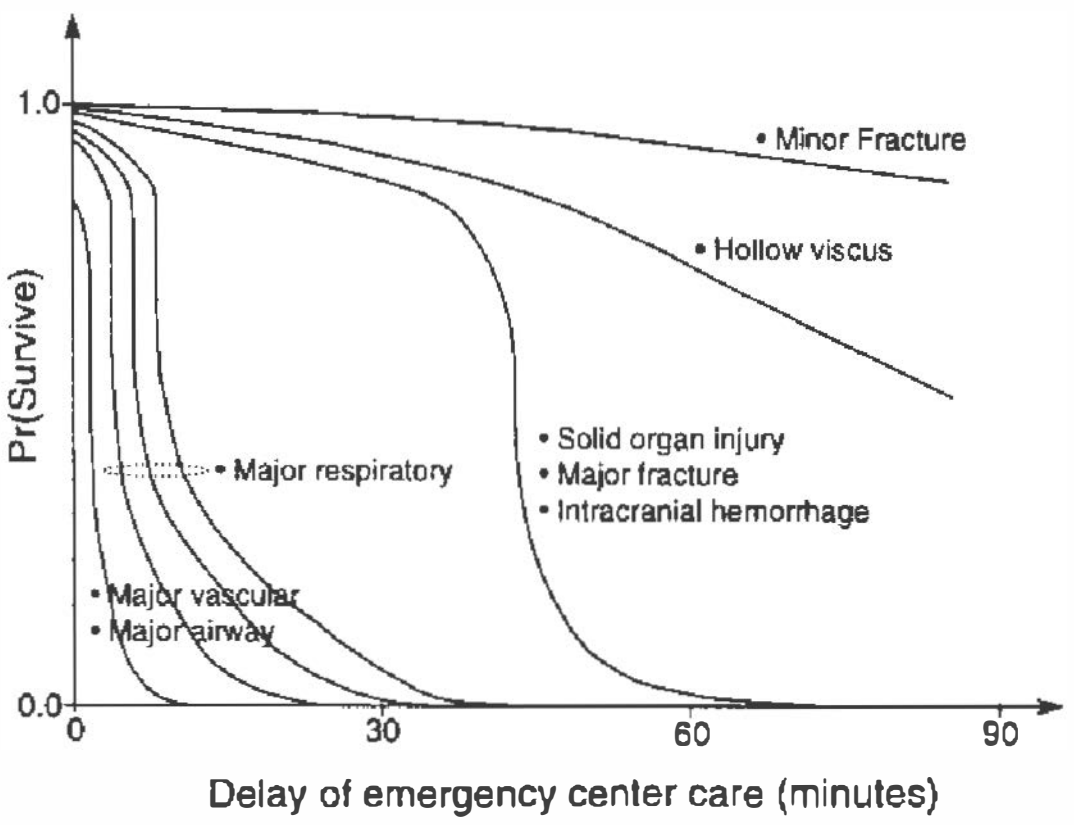

Figure 6: Time-dependent probabilities of survival. These graphs display the expected time-dependent change in the probability of survival as a function of delay to treatment at a trauma center.

### 5.2 Triaging Patients by Cost of Delayed Treatment

For each syndrome, we assess from trauma experts the time-dependent probability of a patient's survival as a function of the delay between the initiation of a destabilizing insult to a patient's physiology and the receipt of attention at a center for treating medical emergencies. Figure 6 displays graphs representing the assessed change in the expected long-term survival of a patient based in the class of injury and delayed treatment for key classes of deranged physiology. Each injury class defines an expected pathological process and default control context. Note that the decreases in survival are nonlinear with delayed treatment. Thus, it is important to include estimates of the duration of a process in the ECDA computation when there is great uncertainty about the time of the trauma.

Additional assessment and modeling can be targeted at conditioning the pathological processes and default control contexts on background physiologies. Consideration of a set of default contexts describing a patient's pre-existing pathophysiologies can be added to the model. For example, the dynamics of the long-term survival with delayed treatment of a severe hemorrhage for a patient with a history of debilitating heart disease would differ from the dynamics for a patient without heart disease.

For any set of findings at the trauma scene, we can employ the Bayesian criticality-assessment model to compute a probability distribution, $p(H_i|E,\xi)$ , over the key major physiological derangements in a patient. Using the probability distribution in conjunction with the time-dependent utility curves allows us to compute for each patient at a site an immediate urgency and an ECDA for alternate transportation actions. Transportation and triage actions to move (or delay moving) a patient to a treatment facility are associated with different expected delays. A pathological process model can provide assistance to decision makers who seek to maximize the response of a trauma system to multiple patients. Alternate transportation and treatment plans can be evaluated in terms of their associated expected costs of delay.

Figure 7 demonstrates inference with the Bayesian model about time-critical physiologies for a set of findings in a motorcycle accident patient. Information input about the patient in the findings worksheet in the lower righthand corner of the screen photo show the signs and symptoms observed about the patient. The upper righthand portion of the photo shows the inferred probability distribution over the pathological processes. The inference shows that the most likely primary pathology in the unconscious patient is a severe permanent injury to the brain. Unfortunately, the outcome of such injuries are not typically sensitive to timely action. Beyond the likelihood of permanent severe brain injury, the system also infers that patient may be threatened by a time-critical pathological problems with significant high ECDA. In particular, there is a 0.1 probability of an intracranial hemorrhage. If this is the case, the patient would benefit

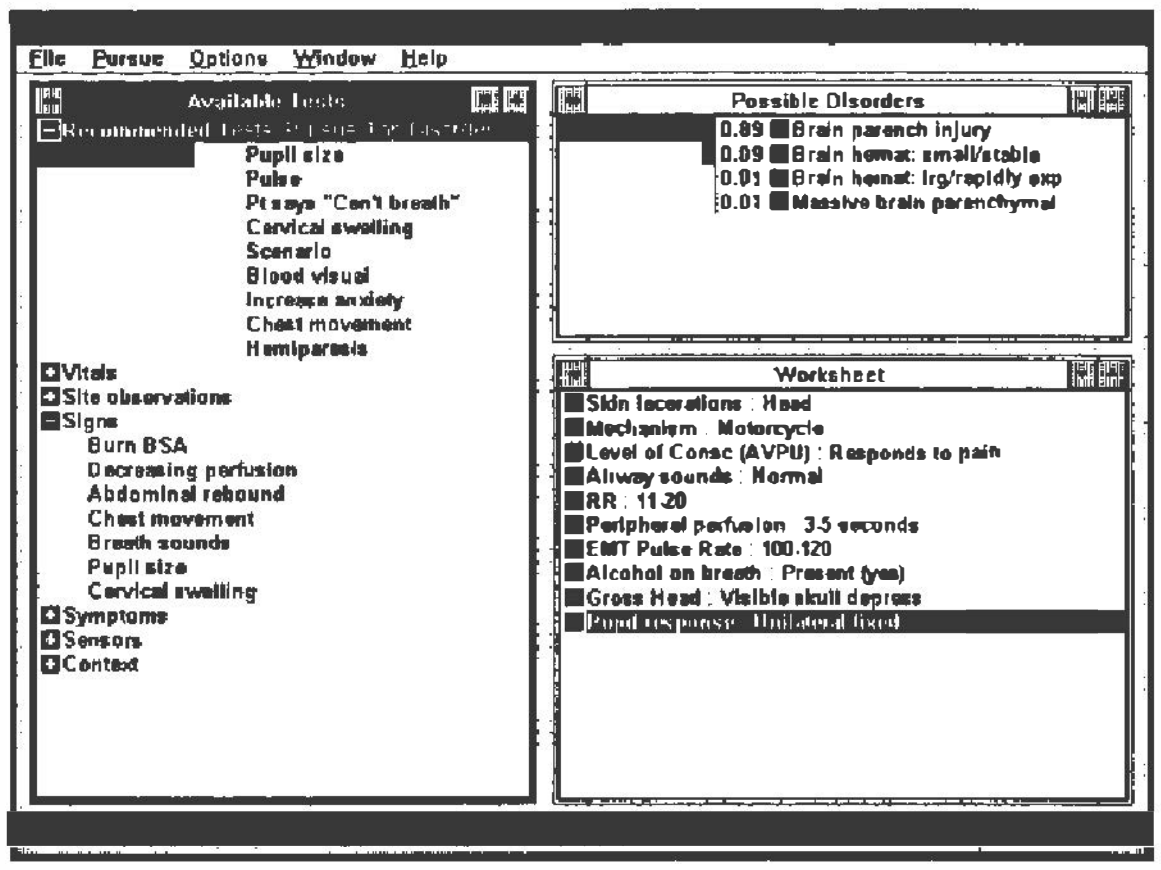

Figure 7: Inference about the status of a victim of a motorcycle accident. Findings are listed in the lower righthand portion of the screen, and the leading processes are ranked by probability in the upper righthand portion of the screen. Recommended findings generated by a value-of-information computation appear on the left side of the screen.

from immediate transportation to a trauma center to minimize the duration of the process.

In a multiple trauma patient setting we can harness the probability distributions over pathological processes and the expected costs of delay to support decisions about the transportation of patients. Transportation actions include decisions to dispatch of assets to as yet unvisited sites, based on telemetered findings about patients at the site. Given multiple transportation assets and treatment facilities with different capacities and capabilities, an automated system can use ECDA computations to evaluate alternate plans.

### 5.3 Modeling the Influence of Uncertainty in Transport Times

In the real-world of emergency medicine, transportation decisions are associated with uncertain delays in moving the patient to a treatment facility. There uncertainty in the amount of time it will take to use alternate routes to transport a patient to a medical center. Travel time depends on such critical variables as the location of the patients, $X$, the location of the treatment facilities, $Y$, and the time of day, $T$, which can indicate the level of traffic on different routes. Decisions about transportation requires consideration of the probability of different transportation times $t$ for each route,

$$= \int_t p(t|X,Y,T)[\max_A \sum_j p(H_j|E,\xi)u(A_i,H_j,t_o) - \max_A \sum_j p(H_j|E,\xi)u(A_i,H_j,t)]dt \quad (5)$$

In a real-world version of a trauma triage systems, dynamically updated traffic reports as well as information about the availability of resources at treatment centers could be used to generate more accurate transportation plans based on the ECDA.

### 5.4 Local Treatment versus Load-and-Go Strategies

In Section 3.1, we discussed the graceful introduction of sets of key intermediate actions to extend the simple time-critical process models. An ongoing area of investigation in the realm of trauma care involves consideration of the merits versus the costs of pausing at a site to initiate treatment of a patient. Immediate treatment can provide early stabilization of pathophysiology but can also incur costs by delaying the transportation of the patient to a trauma center. A formulation of ECDA can be employed for dynamically determining the relative value of local treatment versus immediate transportation of the patient—referred to as a *load-and-go* trauma transportation strategy.

In the general case, we can model the tradeoff between various forms of treating on site and the load-and-go strategy centers on considering local attempts to stabilize a patient by changing the default control context. One approach to assessing the costs and benefits of initial treatment is to consider the influence of an attempt to stabilize the patient as both a means for buying additional time as well as a way to increase the delay to ultimate treatment. That is, we consider the immediate treatment procedures as equivalent to removing some quantity of time from the duration of a pathological process, $t_e(H_j,l,t)$, where $t_e$ refers to the equivalent time removed from the duration of pathologic process $H_i$, by employing local treatment strategy $l$, applied at time $t$. Unfortunately, the commitment to the local therapy will also delay the patient in getting definitive treatment at a trauma center by adding the time required to administer the therapy, $t(l)$. A formulation of ECDA that considers these factors is,

$$\text{ECDA} = \max_A \sum_j p(H_j|E,\xi)u(A_i,H_j,t_o) - \max_A \sum_j p(H_j|E,\xi)u(A_i,H_j,t - t_e(H_i,l,t) + t(l)) \quad (6)$$

## 6 Summary

We have explored the representation of time-dependent decision problems. After reviewing more general models for decision making about time-dependent processes, we discussed pathological processes and simplified models for assessing time-dependent utility and for reasoning about time-critical action. In particular, we investigated models that can be used to diagnosis and to control the duration of pathologic processes. We reviewed the expected cost of delayed action and discussed its use in reasoning about losses with delay. Finally, we focused on an application in the area of trauma care triage and transportation.

## Acknowledgments

We are grateful to Brad Cushing for his assistance with the construction and assessment of Bayesian models for diagnosing classes of pathophysiology. Paul Dagum and Serdar Uckun provided useful feedback on trauma-care issues. The construction of the Bayesian model was supported in part by ARPA in the context of the Trauma Care Information Management Systems project. Bayesian network prototyping was performed with KI software tools.

## References

[1] M. Boddy and T. Dean. Solving time-dependent planning problems. In *Proceedings of the Eleventh IJCAI*. AAAI/International Joint Conferences on Artificial Intelligence, August 1989.

[2] P. Dagum, A. Galper, E. Horvitz, and A. Seiver. Uncertain reasoning and forecasting. *International Journal of Forecasting*, 1995.

[3] R. Dearden and C. Boutilier. Integrating planning and execution in stochastic domains. In *Proceedings of the Tenth Annual Conference on Uncertainty in Artificial Intelligence (UAI-94)*, pages 162–169, Seattle, WA, 1994.

[4] D. Draper, S. Hanks, and D. Weld. A probabilistic model of action for least-commitment planning with information gathering. In *Proceedings of the Tenth Annual Conference on Uncertainty in Artificial Intelligence (UAI-94)*, pages 178–186, Seattle, WA, 1994.

[5] P. Haddawy, A. Doan, and R. Goodwin. Efficient decision-theoretic planning: Techniques and empirical analysis. In *Proceedings of the Eleventh Annual Conference on Uncertainty in Artificial Intelligence (UAI-95)*, pages 229–236, Montreal, Quebec, Canada, 1995.

[6] E. Horvitz. Transmission and display of information: A decision-making perspective. Technical Report Microsoft Technical Report MSR-TR-95-13, Microsoft Research, Microsoft, March 1995.

[7] E. Horvitz and M. Barry. Display of information for time-critical decision making. In *Proceedings of the Eleventh Conference on Uncertainty in Artificial Intelligence*, pages 296–305, Montreal, Canada, August 1995. Morgan Kaufmann, San Francisco, CA.

[8] E. Horvitz, C. Ruokoangas, S. Srinivas, and M. Barry. A decision-theoretic approach to the display of information for time-critical decisions: The Vista project. In *Proceedings of the Conference on Space Operations and Automation and Research, January, 1992*, NASA Johnson Space Center, August 1992. National Aeronautics and Space Administration.

[9] E. Horvitz and G. Rutledge. Time-dependent utility and action under uncertainty. In *Proceedings of Seventh Conference on Uncertainty in Artificial Intelligence, Los Angeles, CA*, pages 151–158. Morgan Kaufmann, San Mateo, CA, July 1991.

[10] E.J. Horvitz. Reasoning under varying and uncertain resource constraints. In *Proceedings AAAI-88 Seventh National Conference on Artificial Intelligence, Minneapolis, MN*, pages 111–116. Morgan Kaufmann, San Mateo, CA, August 1988.

[11] E.J. Horvitz. *Computation and Action Under Bounded Resources.* PhD thesis, Stanford University, 1990.

[12] R. Howard. *Dynamic Programming and Markov Processes.* MIT Press, Cambridge, MA, 1960.

[13] K. Kanazawa, D. Koller, and S. Russell. Stochastic simulation algorithm for dynamic probabilistic networks. In *Proceedings of the Eleventh Annual Conference on Uncertainty in Artificial Intelligence (UAI-95)*, pages 346–351, Montreal, Quebec, Canada, 1995.

[14] M. L. Littman, T. L. Dean, and L. Pack Kaelbling. On the complexity of solving Markov decision problems. In *Proceedings of the Eleventh Annual Conference on Uncertainty in Artificial Intelligence (UAI-95)*, pages 394–402, Montreal, Quebec, Canada, 1995.

[15] R.A. McNutt and S.G. Pauker. Competing rates of risk in a patient with subarachnoid hemorrhage and myocardial infarction: Its now or never. *Medical Decision Making*, 7(4):250–259, 1987.

[16] Mark A. Peot. *Decision-Theoretic Planning.* PhD thesis, Stanford University, 1997.

[17] M.L. Puterman. *Markov Decision Processes: Discrete Stochastic Dynamic Programming.* Wiley, New York, NY, 1994.